\documentclass[letterpaper]{article} 
\usepackage{aaai2026}  
\usepackage{times}  
\usepackage{helvet}  
\usepackage{courier}  
\usepackage[hyphens]{url}  
\usepackage{graphicx} 
\def\UrlFont{\rm}  
\usepackage{natbib}  
\usepackage{caption} 
\usepackage[table,xcdraw]{xcolor}
 \usepackage{booktabs} 
\usepackage{algorithm}
\usepackage{algorithmic}
\usepackage{natbib}
\usepackage{multirow} 
\usepackage{newfloat}
\usepackage{listings}
\usepackage{amsthm,amsmath,amssymb}
\DeclareCaptionStyle{ruled}{labelfont=normalfont,labelsep=colon,strut=off} 
\floatstyle{ruled}
\newfloat{listing}{tb}{lst}{}
\floatname{listing}{Listing}
\title{Rectified Noise: A Generative Model Using Positive-incentive Noise}
\author{
    Zhenyu Gu\textsuperscript{\rm 1,2}\equalcontrib,
    Yanchen Xu\textsuperscript{\rm 1,3}\equalcontrib,
    Sida Huang\textsuperscript{\rm 1,3}\equalcontrib,
    Yubin Guo\textsuperscript{\rm 1,4},
    Hongyuan Zhang\textsuperscript{\rm 1,5}\thanks{Corresponding author.},
}
\affiliations{
    \textsuperscript{\rm 1}Institute of Artificial Intelligence (TeleAI), China Telecom\\
        \textsuperscript{\rm 2}Research Institute of Intelligent Complex Systems, Fudan University\\
    \textsuperscript{\rm 3}School of Artificial Intelligence, OPtics and ElectroNics (iOPEN), Northwestern Polytechnical University\\
    \textsuperscript{\rm 4}University of Science and Technology of China,
    \textsuperscript{\rm 5}The University of Hong Kong\\
}

\usepackage{bibentry}
\begin{document}
\maketitle
\begin{figure*}[!ht]
\centering
\includegraphics[width=\textwidth]{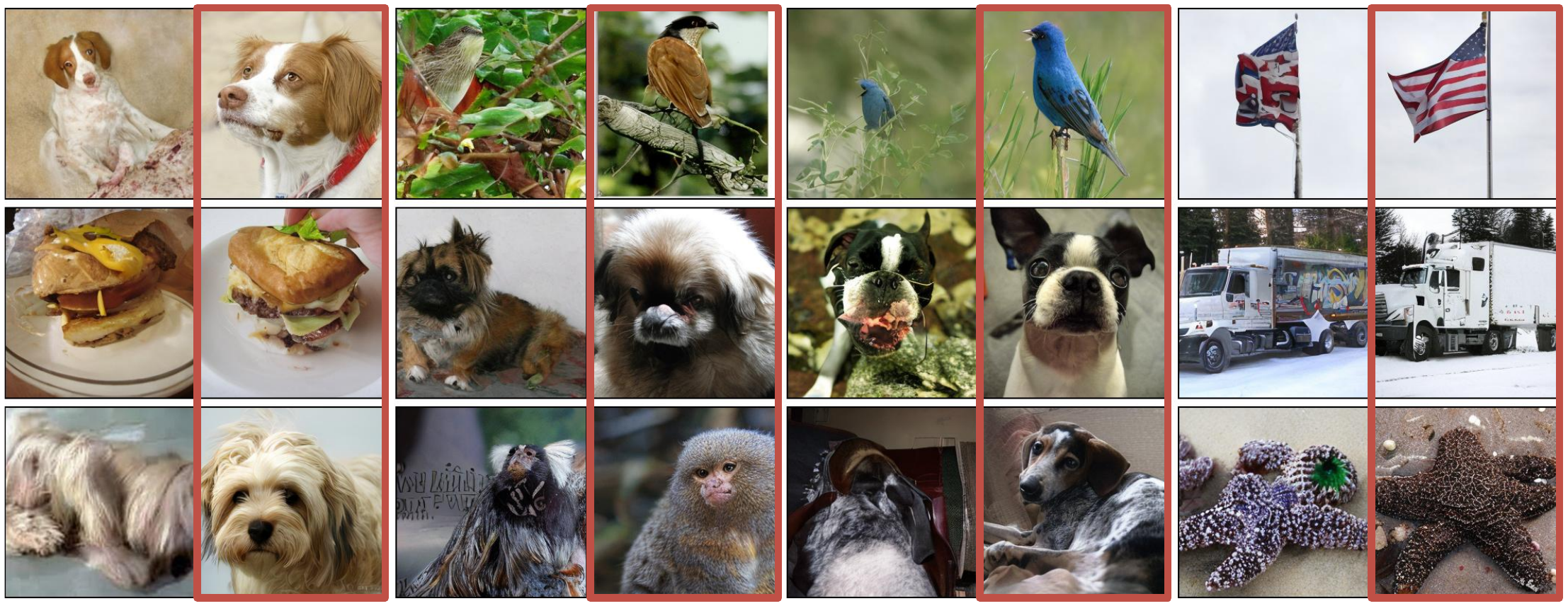} 
\caption{\textbf{Image results of RF models using $\Delta$RN}. Sampling with $\Delta$RN improves natural image generation. The images without red highlight show the generation of the standard RF model. The images outlined in red present the result using $\Delta$RN. Here we show comparisons between images generated by SiT models trained on ImageNet-1k (256 × 256) and SiT models using $\Delta$RN.}
\label{fig2}
\end{figure*}
\begin{abstract}
Rectified Flow (RF) has been widely used as an effective generative model. Although RF is primarily based on probability flow Ordinary Differential Equations (ODE), recent studies have shown that injecting noise through reverse-time Stochastic Differential Equations (SDE) for sampling can achieve superior generative performance. Inspired by Positive-incentive Noise ($\pi$-noise), we propose an innovative generative algorithm to train $\pi$-noise generators, namely Rectified Noise ($\Delta$RN), which improves the generative performance by injecting $\pi$-noise into the velocity field of pre-trained RF models. After introducing the Rectified Noise pipeline, pre-trained RF models can be efficiently transformed into $\pi$-noise generators. We validate Rectified Noise by conducting extensive experiments across various model architectures on different datasets. Notably, we find that: (1) RF models using Rectified Noise reduce FID from \textbf{10.16 to 9.05} on ImageNet-1k. (2) The models of $\pi$-noise generators achieve improved performance with only \textbf{0.39\%} additional training parameters. Code is available here: https://github.com/simulateuser538/Rectified-Noise
\end{abstract}

%

\section{Introduction}
Flow Matching (FM) \cite{lipman2022flow,albergo2022building,liu2022flow} for generative models trains continuous normalizing flows \cite{papamakarios2021normalizing} by regressing ideal probability flow fields that connect a base distribution to the data distribution. FM models show superior performance and has seen widespread adoption in modern generative modeling \cite{esser2024scaling,polyak2024movie,fu2025objectaveditobjectlevelaudiovisualediting}. Rectified Flow (RF) \cite{liu2022flow} is a specific kind of FM that simplifies the training objective by prescribing a straight-line path between the source and target distributions. Different from diffusion models relying on reverse-time Stochastic Differential Equation (SDE) \cite{song2020score} , RF directly learns the velocity field that transforms an analytic distribution into the target data distribution without introducing additional  stochasticity. Training RF models through a simple regression-based objective enables more stable and efficient training.

However, recent studies have demonstrated that introducing stochastic noise to a pretrained RF model during sampling, specifically via a reverse-time SDE \cite{ma2024sit} can improve performance metrics like Fréchet Inception Distance (FID) \cite{heusel2017gans}. This phenomenon inspires us to investigate: 

\textbf{(1) What kind of stochasitic noise can lead to performance gain for RF?}

\textbf{(2) How to introduce the beneficial noise to RF?}

Inspired by Positive-incentive Noise ($\pi$-noise) \cite{li2022positive,zhang2025variational,huang2025enhance}, it may be a reliable scheme to apply $\pi$-noise framework to learn the beneficial noise for improving the performance of RF models. Specifically, the goal of $\pi$-noise is to find the beneficial noise by maximizing the mutual information \cite{shannon1948mathematical} 
between the task and noise. The existing works have shown that $\pi$-noise can be used to effectively enhance both the classical neural networks \cite{zhang2025variational,zhang2024dataaugmentationcontrastivelearning,huang2025learnbeneficialnoisegraph,jiang2025mixturenoisepretrainedmodelbased} and vision-language models \cite{huang2025enhance,huang2025nfigmultiscaleautoregressiveimage,wang2025advcpgcustomizedportraitgeneration}. The success of applying $\pi$-noise to enhance model performance adds to the rationality to apply $\pi$-noise to generative models as stated above. In this paper, we propose Rectified Noise ($\Delta$ RN), a novel framework that leverages learned $\pi$-noise to enhance the performance of RF models. The contributions can be briefly summarized as follows:
\begin{itemize}
    \item Under the $\pi$-noise framework, we measure the complexity of RF by designing an auxiliary Gaussian distribution related to RF loss. The auxiliary Gaussian variable connects RF and information entropy.
    \item Motivated by the connection between $\pi$-noise and RF, we propose a $\pi$-noise generator to automatically learn the additional noise component when solving the velocity in RF. We further design a framework to convert pre-trained RF models into $\pi$-noise generators. 
    \item Experiments on multiple datasets including
    ImageNet, AFHQ and CelebA-HQ validate the effectiveness of our proposed $\Delta$RN. Our experiments show this framework achieves performance improvements while maintaining computational efficiency. $\Delta$RN enhances RF performance across all datasets, reducing FID by up to \textbf{1.11} on ImageNet,\textbf{ 1.89} on AFHQ and \textbf{3.52} on CelebA-HQ.
\end{itemize}
\section{Related Work}

In this section, we will discuss the related work about generative models, Scalable Interpolant Transformers (SiT)\cite{ma2024sit}, and $\pi$-noise, respectively. 

\subsection{Generative Model}
Diffusion models \cite{sohl2015deep,song2019generative,ho2020denoising} have been developed into a highly successful framework for generative modeling. These models progressively add noise to clean data and train a neural network to reverse this process.
Flow Matching (FM) \cite{lipman2022flow,albergo2022building,liu2022flow} methods extend this framework trains continuous normalizing flows \cite{papamakarios2021normalizing} by regressing ideal probability flow fields that connect a base distribution to the data distribution.

Rectified Flow (RF) offers an efficient alternative in generative modeling. It directly parameterizes continuous-time transport, greatly reducing sampling steps. Unlike diffusion models using separate score estimation \cite{song2019generative,vahdat2021score} and noise, RF learns a clear, deterministic map between data and latent distributions with probability flow ODEs \cite{chen2018neural,papamakarios2021normalizing,zheng2023improved}. This direct method simplifies generation by avoiding noisy and iterative steps and usually improves training. RF uses a straight-line sample pairing strategy to define a simple and consistent path between two distributions and uses reflow to cut the high computational cost of diffusion sampling while keeping high-quality image generation with fewer steps.
\subsection{Scalable Interpolant Transformers}
SiT represents a novel family of generative models, building upon the foundation of Diffusion Transformers (DiT) \cite{peebles2023scalable}. SiT is an extension of Vision Transformer (ViT) \cite{dosovitskiy2020image} that operates within the stochastic interpolant framework \cite{albergo2023stochastic}. Its primary contributions include a systematic exploration of the design space for generative models, encompassing aspects like time discretization, model prediction, interpolants and samplers. This systematic approach has not only facilitated a modular study of each component but also led to the discovery of optimal practices for enhancing generation performance. SiT also explores the performance gains brought by the interpolant framework under Classifier-Free Guidance (CFG) \cite{ho2022classifier}. 

Furthermore, a key finding of SiT's research pertains to the use of reverse-time SDE for sampling of flow matching models. Using  reverse-time SDE sampling \cite{song2020score} often leads to better performance compared to probability flow ODE sampling \cite{song2019generative}.
\subsection{Positive-incentive Noise}
$\pi$-noise introduces an information-theoretic framework to formally claim that noise may not always be harmful. 
$\pi$-noise can be seen as a type of information gain brought by noise. This approach proposes learning the $\pi$-noise by maximizing the mutual information between the task and the noise. To optimize the intractable loss of $\pi$-noise, VPN \cite{zhang2025variational} proposed to optimize its variational bound and PiNI \cite{huang2025enhance} extended it to vision-language models. With the variational inference, a VPN generator is designed for enhancing base models and simplifying the inference without changing the architecture of base models.

\section{Preliminaries}

In this section, we provide a brief overview of RF model from the perspective of stochastic interpolants \cite{albergo2023stochastic} and revisit the $\pi$-noise framework. 
\begin{figure*}[t]
\centering
\includegraphics[width=0.93\textwidth]{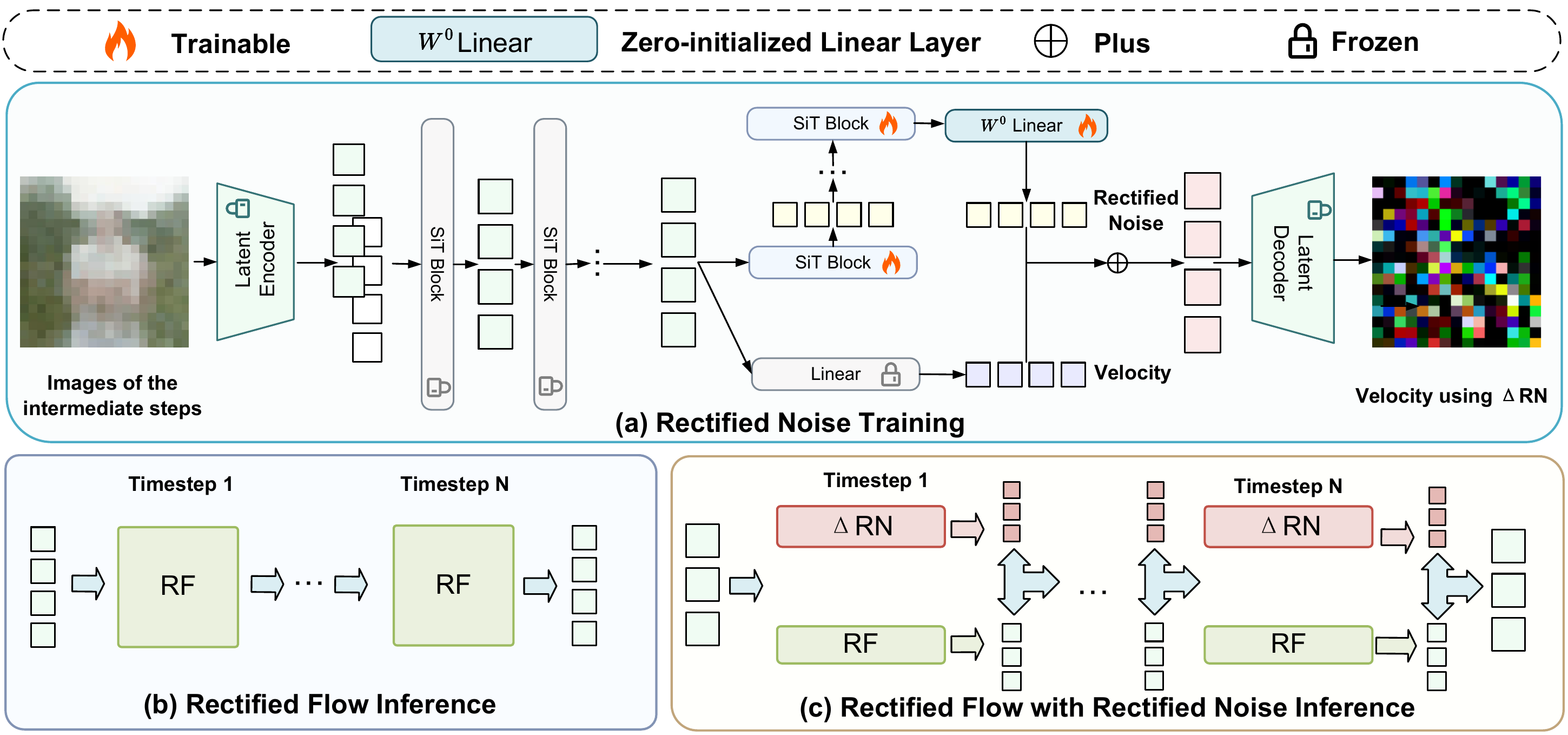}
\caption{\textbf{Overview of Rectified Noise pipeline.} \textbf{(a)} The Rectified Noise model inherits pre-trained knowledge from a foundation model (RF) through parameter freezing.  Additional and trainable SiT blocks are integrated to predict $\pi$-noise. \textbf{(b)} Inference of traditional RF models. \textbf{(c)} Inference with the Rectified Noise involves adding $\pi$-noise to the predicted velocity field.}
\label{pipe1}
\end{figure*}
\subsection{Rectified Flow}
RF aims to learn a transport map from a standard Gaussian noise distribution $\mathcal{N}(0, \textit{I})$ to an arbitrary distribution $q(\mathbf{x_*})$ defined on the reals. Specifically, the goal is to gradually transform an initial noise sample $\boldsymbol{\epsilon} \sim \mathcal{N}(0, \textit{I})$ over time into data $\mathbf{x_*}\sim q(\mathbf{x_*})$ for the generating task. Stochastic interpolants  define this transformation as a time-dependent stochastic process, which can be summarized as 
\begin{equation}
    \mathbf{{x}_t}=t\mathbf{{x}_*}+(1-t)\boldsymbol{\epsilon},
\end{equation}

RF models interpolate between noise and data over a finite time interval defined on $t \in [0, 1]$.
Sampling from these models  can be achieved via a probability flow ordinary differential equation (\textbf{Probability Flow ODE}) with a velocity field 
\begin{equation}
\dot{\mathbf{x}}_t=\mathbf{v}(\mathbf{x}_t,t),
\end{equation}
where the velocity field $\mathbf{v}(\mathbf{x}_t,t)$ is given by the conditional expectation
\begin{equation}
\begin{aligned}
\mathbf{v}(\mathbf{x},t) & =\mathbb{E}[\dot{\mathbf{x}}_t|\mathbf{x}_t=\mathbf{x}], \\
\end{aligned}
\end{equation}
where $\mathbf{v}(\mathbf{x}_t,t)$ signifies the expected direction of all transport paths between the noise and $p(\mathbf{x})$ that cross through $\mathbf{x}_t$ at time $t$. We can estimate $\mathbf{v}(\mathbf{x}_t,t)$ via the loss
\begin{equation}
    \mathcal{L}_\mathbf{velocity}(\theta):=\mathbb{E}_{\mathbf{x_*},\boldsymbol{\epsilon},t}\left[\left\|\mathbf{v}_\theta(\mathbf{x}_t,t)-\mathbf{x_*}+\boldsymbol{\epsilon}\right\|^2\right].
\end{equation}

\textbf{Reverse-time SDE} \cite{tzen2019theoretical,vargas2021solving,de2021diffusion,song2020score} offers an alternative sampling method for flow matching, which can be expressed as
\begin{equation}
    \mathrm{d}\mathbf{x}_t=\mathbf{v}(\mathbf{x}_t,t)\mathrm{d}t-\frac{1}{2}w_t\mathbf{s}(\mathbf{x}_t,t)\mathrm{d}t+\sqrt{w_t}\mathrm{d}\bar{\mathbf{W}}_t,
\end{equation}
where $\bar{W}_t$ is a reverse-time Wiener process, $w_t > 0$ denotes an arbitrary time-dependent diffusion coefficient, $\mathbf{v}(\mathbf{x}_t,t)$ refers to the velocity and $\mathbf{s}(\mathbf{x}_t, t) = \nabla \log q_t(\mathbf{x}_t)$ is identified as the score, which is also determined by a conditional expectation
\begin{equation}
    \begin{aligned}
\mathbf{s}(\mathbf{x}_t,t) & =-\frac{\mathbb{E}[\boldsymbol{\epsilon}\mid\mathbf{x}_t=\mathbf{x}]}{1-t} \\
 & =\frac{\mathbf{x}_*-\mathbf{x}_t}{(1-t)t^2}.
\end{aligned}
\end{equation}

Originally, {reverse-time SDE} were used in score-based diffusion models, where the diffusion coefficient $w_t$ typically depended on the forward SDE \cite{song2020score,chen2023importance,singhal2023diffuse}. The stochastic interpolant framework provides greater flexibility by decoupling the formulation of $\mathbf{x}_t$ from the forward SDE, which allows for a wider choice of $w_t$: any $w_t \ge 0$ can be used. It's noteworthy that the choice of $w_t$ can be made after training, as it does not impact the velocity $\mathbf{v}(\mathbf{x}, t)$ or the score $\mathbf{s}(\mathbf{x}, t)$.
\subsection{Formulation of $\pi$-Noise}
$\pi$-noise is primarily studied from an information-theoretic perspective. The goal of $\pi$-noise is to find the beneficial noise by maximizing the mutual information $\max_{\mathcal{E}} \operatorname{MI}(\mathcal{T},\mathcal{E})$ between the task and noise. The principle of learning $\pi$-noise is formulated as
\begin{equation}
\label{information}
\begin{aligned}
\max_{\mathcal{E}} \operatorname{MI}(\mathcal{T},\mathcal{E}) &= H(\mathcal{T}) - H(\mathcal{T}|\mathcal{E})
\Leftrightarrow \max_{\mathcal{E}} - H(\mathcal{T}|\mathcal{E}),
\end{aligned}
\end{equation}
where $H(\cdot)$ represents the information entropy. 

The task entropy  $H(\mathcal{T})$ is the core of this framework. To measure the difficulty of a RF learning task for a given dataset $\mathcal{D}$ sampled from $p(\mathbf{x})$, it is essential to properly define a random variable for the task. Building on this, we can further derive the expression for $H(\mathcal{T}|\mathcal{E})$.
\section{Rectified Noise}
In this section, we elaborate on the proposed Rectified Noise, a novel approach for injecting $\pi$-noise into the velocity of pre-trained RF models. We first demonstrate how to define the task entropy of RF models. Subsequently, we demonstrate how to learn $\pi$-noise distribution under this definition. Finally, we propose two optimization strategies for the objective of learning
$\pi$-noise and design Rectified Noise pipeline to transform RF models into $\pi$-noise generators.
\subsection{Formulate Task Entropy via RF Loss}
Measuring the learning complexity of RF models across diverse datasets is a challenge problem. Therefore, we concentrate on measuring this complexity by defining the task entropy on a given dataset.

Considering a given distribution $q(\mathbf{x_*})$, the value of loss $\mathcal{L}_\mathbf{velocity}({\psi^*})$ can serve as a measure of generation task difficulty for RF, where ${\psi^*}$ represents the optimal parameters of the neural network model. To simplify the derivation,
let $\mathbf{x} = (\mathbf{x_*}, \mathbf{x_0})\sim p(\mathbf{x})$ where $\mathbf{x_*}\sim q(\mathbf{x_*})$ and $\mathbf{x_0}\sim \mathcal{N}(0,\textit{I})$  and let
\begin{equation}
\begin{aligned}
    \mathcal{L}_\mathbf{velocity}({\psi^*})&= \mathbb{E}_{\mathbf{x},t}  \mathcal{L}_(\mathbf{x},t;{\psi^*})\\&=\mathbb{E}_{\mathbf{x},t}  \left[\left\|\mathbf{v}_{{\psi^*}}(\mathbf{x}_t,t)-\mathbf{x}_*+\mathbf{x_0}\right\|^2\right].
    \end{aligned}
\end{equation}
The smaller the value of $\mathcal{L}(\mathbf{x})$ is, the easier it is for the neural network to fit the velocity field generated by interpolating the data pair $\mathbf{x} = (\mathbf{x_*}, \mathbf{x_0})$ and vice versa.

To bridge the framework of $\pi$-noise and the complexity metric $\mathcal{L}(\mathbf{x},t; {\psi^*})$, we introduce an auxiliary random variable $\alpha$ , satisfying 
\begin{equation}
\alpha |\mathbf{x},t \sim \mathcal{N}(0,\exp(\mathcal{L}(\mathbf{x},t;{\psi^*}))).
\end{equation}
The information entropy of the auxiliary distribution $p(\alpha|\mathbf{x})$ reflects the difficulty for the corresponding generative model parameterized by ${\psi^*}$. Consequently, for a given distribution $q(\mathbf{x_*})$, the task entropy of the
generation task $\mathcal{T}$ can be written as 
\begin{equation}
\begin{aligned}
     H(\mathcal{T})&= \mathbb{E}_{\mathbf{x},t}  H(p(\alpha|\mathbf{x},t)) \\
    &= \frac{1}{2} \mathbb{E}_{\mathbf{x},t}  \mathcal{L}(\mathbf{x},t; {\psi^*})+\frac{1}{2} \ln(2\pi e).
\end{aligned}
\end{equation}
\subsection{Inject $\pi$-Noise to RF Model}
Now, we show how we learn $\pi$-noise for a RF model by uncovering the connection between $\pi$-noise and RF. With the definition of $H(\mathcal{T})$ given in the previous subsection, the mutual information between the generation task $\mathcal{T}$ and the noise $\mathcal{E}$ can be calculated as
\begin{equation}
\begin{aligned}
    \operatorname{MI}(\mathcal{T}|\mathcal{E}) &= \mathbb{E}_{\mathbf{x},t}  \int p(\alpha, \boldsymbol{\epsilon}|\mathbf{x},t) \log \frac{p(\alpha, \boldsymbol{\epsilon}|\mathbf{x},t)}{p(\alpha|\mathbf{x},t) p(\boldsymbol{\epsilon}|\mathbf{x},t)} d\alpha d\boldsymbol{\epsilon}   \\
    &=\int p(\alpha, \boldsymbol{\epsilon},\mathbf{x},t) \log \frac{p(\alpha, \boldsymbol{\epsilon}|\mathbf{x},t)}{p(\alpha|\mathbf{x},t) p(\boldsymbol{\epsilon}|\mathbf{x},t)} d\mathbf{x} d\boldsymbol{\epsilon}d\alpha dt .
\end{aligned}
\end{equation}
Similarly, the formulation of the conditional entropy can be written as $H(\mathcal{T} | \mathcal{E})=$
\begin{equation}
\begin{aligned}
 -\int p(\alpha|\mathbf{x}, \boldsymbol{\epsilon},t) p(\boldsymbol{\epsilon}|\mathbf{x},t) p(\mathbf{x},t) \log p(\alpha|\mathbf{x}, \boldsymbol{\epsilon},t) d\mathbf{x} d\boldsymbol{\epsilon}d\alpha dt.
\end{aligned}
\end{equation}

As shown in Eq. ($\ref{information}$), maximizing mutual information can be achieved by minimizing conditional entropy. We consider the dataset $\mathcal{D}$ sampling from the joint distribution of $(\mathbf{x},t)$. Using Monte Carlo method, $H(\mathcal{T}|\mathcal{E})$ can be approximately written as $H(\mathcal{T}|\mathcal{E}) \approx$
\begin{equation}
\begin{aligned}
 -\frac{1}{|\mathcal{D}|} \sum_{(\mathbf{x},t) \in \mathcal{D}} \int p(\alpha|\mathbf{x},\boldsymbol{\epsilon}, t) p(\boldsymbol{\epsilon}|\mathbf{x},t) \log p(\alpha|\mathbf{x}, \boldsymbol{\epsilon},t) d\alpha d\boldsymbol{\epsilon}.
\end{aligned}
\end{equation}
The formulation on the right involves two probabilities: $p(\alpha|\mathbf{x}, \boldsymbol{\epsilon},t)$ and $p(\boldsymbol{\epsilon}|\mathbf{x},t)$. We learn $p(\boldsymbol{\epsilon}|\mathbf{x},t)$ as the distribution of $\pi$-noise with learnable parameters, making it essential to accurately model $p(\alpha|\mathbf{x}, \boldsymbol{\epsilon},t)$. Then, we can define the auxiliary distribution with $\boldsymbol{\epsilon}$ as
\begin{equation}
\alpha |\mathbf{x},\boldsymbol{\epsilon},t \sim \mathcal{N}(0,\exp(\mathcal{L}(\mathbf{x},\boldsymbol{\epsilon},t;{\psi^*}))),
\end{equation}
where 
\begin{equation}
\mathcal{L}(\mathbf{x},\boldsymbol{\epsilon},t,{\psi^*})=\left\|\mathbf{v}_{{\psi^*}}+\boldsymbol{\epsilon}(\mathbf{x}_t,t)-\mathbf{x}_*+\mathbf{x_0}\right\|^2.
\end{equation}
The above formula is equivalent to injecting noise into the velocity field of a pre-trained RF model.

It should be pointed out that the optimization objective will be completely equivalent to RF model if we employ a point estimation of $p(\boldsymbol \epsilon | \mathbf x, t)$ 
for a given $(\mathbf{x}, t)$, i.e.,
\begin{equation}
p(\boldsymbol{\epsilon}|\mathbf{x}, t) \rightarrow \delta(\boldsymbol{\epsilon}),
\end{equation}
where $\delta(\boldsymbol{\epsilon})$ denotes the Dirac delta function, which satisfies
\begin{equation}
\delta(\boldsymbol{\epsilon}) =
\begin{cases}
    \infty & \text{if } \boldsymbol{\epsilon} = 0 \\
    0 & \text{if } \boldsymbol{\epsilon} \neq 0
\end{cases}  \quad \rm and
\int_{-\infty}^{\infty} \delta(\boldsymbol{\epsilon})d\boldsymbol{\epsilon} =1.
\end{equation}
With the point estimation, $H(\mathcal{T} | \mathcal{E})$ can be simplified as 
\begin{equation}
\begin{aligned}
&H(\mathcal{T}|\mathcal{E})\\
&\approx -\frac{1}{|\mathcal{D}|} \sum_{(\mathbf{x},t) \in \mathcal{D}} \int p(\alpha|\mathbf{x},\boldsymbol{\epsilon}=\mathbf{0}, t)\log p(\alpha|\mathbf{x}, \boldsymbol{\epsilon}=\mathbf{0},t) d\alpha \\
& =- \frac{1}{2} \ln(2\pi e) -\mathbb{E}_{\mathbf{x},t}  \mathcal{L}(\mathbf{x},\boldsymbol{\epsilon}=\mathbf{0},t; {\psi^*}),
\end{aligned}\end{equation}
which is equivalent to the loss of RF models. 
The estimation indicates that $\pi$-noise always keeps 0 in RF models. 
Accordingly, we can \textbf{learn the $\pi$-noise, instead of simply estimating it}, and get $\Delta$RN as 
\begin{equation}
\begin{aligned}
&\max -\frac{1}{|\mathcal{D}|} \sum_{(\mathbf{x},t) \in \mathcal{D}} \int p(\alpha, \boldsymbol{\epsilon}|\mathbf{x}, t)\log p(\alpha|\mathbf{x}, \boldsymbol{\epsilon},t) d\alpha \\
\Leftrightarrow & \max -\frac{1}{|\mathcal{D}|} \mathbb{E}_{\boldsymbol{\epsilon}}\sum_{(\mathbf{x},t) \in \mathcal{D}} \int p(\alpha |\mathbf{x}, \boldsymbol{\epsilon}, t)\log p(\alpha|\mathbf{x}, \boldsymbol{\epsilon},t) d\alpha \\
\Leftrightarrow & \max \mathbb{E}_{\boldsymbol{\epsilon}, \mathbf{x},t} \mathcal{L}(\mathbf{x},\boldsymbol{\epsilon},t; {\psi^*}) .
\end{aligned}
\end{equation}

Given a specific RF model (with ${{\psi^*}}$ being optimal parameters), $\boldsymbol{\epsilon}$ is  determined by $\mathbf{x}$ and $t$. We use a neural network parameterized by $\theta$ to represent $\boldsymbol{\epsilon}$, denoted as $\boldsymbol{\epsilon}_\theta$. Leveraging the aforementioned derivation, the optimization objective for the $\pi$-noise can be equivalently formulated as
\begin{equation}
\begin{aligned}
\max_{\mathcal{E}} \operatorname{MI}(\mathcal{T},\mathcal{E})  \Leftrightarrow \max_{\theta}\mathbb{E}_{\mathbf{x},t, \boldsymbol{\epsilon} \sim \boldsymbol{\epsilon}_\theta} \mathcal{L}(\mathbf{x},\boldsymbol{\epsilon},t; {\psi^*}).   
\end{aligned}
\end{equation}
\subsection{Optimization Strategies for $\pi$-Noise}
To optimize the objective $\max_{\theta}\mathbb{E}_{\mathbf{x},t, \boldsymbol{\epsilon} \sim \boldsymbol{\epsilon}_\theta}  \mathcal{L}(\mathbf{x},\boldsymbol{\epsilon}_\theta,t;{\psi^*})$ proposed in the previous section, we discuss the two cases in applications: (1) Train $\Delta$RN and RF simultaneously (i.e., learning $\theta$ and $\psi$); (2) Only train $\Delta$RN for a pre-trained RF model (i.e., learn $\theta$ with frozen ${\psi^*}$).

\subsubsection{{4.3.1 Optimize both $\theta$ and $\psi$}}
\quad\\
$\theta$ and $\psi$ are optimized simultaneously. This can be achieved by adjusting the assumption on the distribution of $\boldsymbol{\epsilon}$, thereby unifying parameters $\theta$ and $\psi$ into a single neural network.

To facilitate predictions by the neural network model, we select three common reparameterizable distributions as the assumed distributions for $\pi$-noise:
\begin{itemize}
    \item \textbf{Gaussian Distribution} 
\begin{equation}
    \mathbf{z} = \boldsymbol{\mu} + \boldsymbol{\sigma} \odot \boldsymbol{\epsilon},\quad \boldsymbol{\epsilon} \sim \mathcal{N}(0, \textit{I})
\end{equation}

    \item \textbf{Gumbel Distribution} 
\begin{equation}
  \mathbf{z} = \boldsymbol{\mu} - \boldsymbol{\beta} \odot\log(-\log(\boldsymbol{\epsilon})),\quad{\epsilon_i}\sim U(0, {1}) 
\end{equation}
    
    \item \textbf{Uniform Distribution} 
\begin{equation}
    \mathbf{z}  =  \mathbf{a}  + ( \mathbf{b}- \mathbf{a} ) \odot \boldsymbol{\epsilon},\quad{\epsilon_i}\sim U(0, {1}) 
\end{equation}
    
\end{itemize}
where $\odot$ is Hadamard product operator and $U(0,1)$  represents a uniform distribution over the interval [0,1].

Taking the Gaussian distribution as an example, we will explain how to achieve the unification of $\theta$ and ${\psi^*}$. Leveraging the reparameterization trick, the initial optimization objective is reformulated as
\begin{equation}
\begin{aligned}
&\mathbb{E}_{\mathbf{x},t, \boldsymbol{\epsilon} \sim \boldsymbol{\epsilon}_\theta}  \mathcal{L}(\mathbf{x},\boldsymbol{\epsilon}_\theta,t;{\psi^*})
\\
=&\mathbb{E}_{\mathbf{x},t, \boldsymbol{\epsilon} \sim \boldsymbol{\epsilon}_\theta} \left\|\mathbf{v}_{{\psi^*}}+\boldsymbol{\epsilon}_\theta(\mathbf{x}_t,t)-\mathbf{x}_*+\mathbf{x_0}\right\|^2\\
=&\mathbb{E}_{\mathbf{x},t, \boldsymbol{\epsilon} \sim \boldsymbol{\epsilon}_\theta} \left\|\mathbf{v}_{{\psi^*}}+\boldsymbol{\mu}_\theta(\mathbf{x}_t,t)+\boldsymbol{\sigma}_\theta(\mathbf{x}_t,t)\odot\boldsymbol{\epsilon}-\mathbf{x}_*+\mathbf{x_0}\right\|^2,\\
\end{aligned}
\end{equation}
where $\boldsymbol{\epsilon} \sim \mathcal{N}(0, \textit{I})$. Since $\mathbf{v}_{{\psi^*}}$ can essentially be regarded as a constant determined by $\mathbf{x}_t$ and $t$, $\boldsymbol{\hat{\mu}}_\theta = \mathbf{v}_{{\psi^*}}(\mathbf{x}_t, t) + \boldsymbol{\mu}_\theta(\mathbf{x}_t, t)$ can be treated as a single entity and predicted by a single neural network. Algorithm \ref{alg:algorithm1} illustrates the implementation of an arbitrary batch step. Optimization of Gumbel distribution and uniform distribution is similar to the optimization of Gaussian distribution.

\begin{algorithm}[t]
\caption{Pseudo code for batch step of optimizing $\theta$ without pre-trained RF model }
\label{alg:algorithm1}
\textbf{Input}: A model $\boldsymbol{\epsilon}_\theta=\boldsymbol{\mu}_\theta+\boldsymbol{\sigma}_\theta \odot \boldsymbol{\epsilon}$, batch of $N$ flow examples $F=\{(\mathbf{x_i},\mathbf{y_i})\}$ where $(\mathbf{x_i},\mathbf{y_i})\sim p(\mathbf{x})$, learning rate $\beta$  \\
\textbf{Output}: Updated parameters $\theta$ 
\begin{algorithmic}[1] 
\STATE Let $L(\theta)=0$.
\STATE \textbf{for} $i$ in range($N$) \textbf{do}
\STATE \qquad  $t\sim U(0,1),\mathbf{x}_t=t\mathbf{x_i}+(1-t)\mathbf{y_i}$
\STATE \qquad  Sample $\boldsymbol{\epsilon}\sim\mathcal{N}(0,\textit{I})$
\STATE\qquad\# \textit{Depending on the noise assumption,} 
\STATE\qquad\# \textit{the distribution of $\boldsymbol{\epsilon}$ can be adjusted.} 
\STATE \qquad  $\mathbf{\hat{v}}=\boldsymbol{\mu}_\theta(\mathbf{x}_t,t)+\boldsymbol{\sigma}_\theta(\boldsymbol{x_t},t)\odot\boldsymbol{\epsilon}$, $\mathbf{v}=\mathbf{x_i}-\mathbf{y_i}$
\STATE \qquad $L(\theta)+=\|\mathbf{\hat{v}}-\mathbf{{v}}\|^2$
\STATE \textbf{end for}
\STATE $\theta\leftarrow\theta-\frac{\beta}{N}\nabla_\theta L(\theta)$
\end{algorithmic}
\end{algorithm}

\subsubsection{{4.3.2 Optimize $\theta$ with frozen ${\psi^*}$}}
\quad\\
Initially, we learn the RF parameters ${\psi^*}$. Following this, $\theta$ is optimized while ${\psi^*}$ is frozen. By fine-tuning the pre-trained RF neural network, we can more efficiently learn the parameter $\theta$.
\begin{algorithm}[t]
\caption{Pseudo code for batch step of optimizing $\theta$ with pre-trained RF model}
\label{alg:algorithm2}
\textbf{Input}:A pre-trained RF model $\mathbf{v}_{\psi^*}=\boldsymbol{f}_{\psi^*} \circ \mathbf{s}_{\psi^*}$
($\boldsymbol{f}_{\psi^*}$ is linear layer and $\mathbf{s}_{\psi^*}$ are sit blocks function), a model $\boldsymbol{\epsilon}_\theta(\mathbf{s}_{\psi^*}(\mathbf{x}_t,t))=\boldsymbol{\mu}_\theta+\boldsymbol{\sigma}_\theta\odot\boldsymbol{\epsilon}$ , batch of $N$ flow examples $F=\{(\mathbf{x}_i,\mathbf{y_i})\}$ where $(\mathbf{x_i},\mathbf{y_i})\sim p(\mathbf{x})$, learning rate $\beta$  \\
\textbf{Output}: Updated parameters $\theta$ 
\begin{algorithmic}[1] 
\STATE Let $L(\theta)=0$
\STATE \textbf{for} $i$ in range($N$) \textbf{do}
\STATE \qquad  $t\sim U(0,1), \mathbf{x}_t=t\mathbf{x_i}+(1-t)\mathbf{y_i}$ 
\STATE \qquad  $\mathbf{\hat{s}}=\mathbf{s}_{\psi^*}(\mathbf{x}_t,t)$
\STATE \qquad  Sample $\boldsymbol{\epsilon}\sim\mathcal{N}(0,\textit{I})$\\
\STATE\qquad\# \textit{Depending on the noise assumption,} 
\STATE\qquad\# \textit{the distribution of $\boldsymbol{\epsilon}$ can be adjusted.} 
\STATE \qquad  $\mathbf{\hat{v}}=\boldsymbol{f}_{\psi^*}(\mathbf{\hat{s}})+\boldsymbol{
\mu}_\theta(\mathbf{\hat{s}})+\boldsymbol{\sigma}_\theta(\mathbf{\hat{s}})\odot\boldsymbol{\epsilon}$, $\mathbf{v}=\mathbf{x_i}-\mathbf{y_i}$
\STATE \qquad $L(\theta)+=\|\mathbf{\hat{v}}-\mathbf{{v}}\|^2$
\STATE \textbf{end for}
\STATE $\theta\leftarrow\theta-\frac{\beta}{N}\nabla_\theta L(\theta)$
\end{algorithmic}
\end{algorithm}
\begin{table*}[t]
\centering
\begin{tabular}{crcccccccc}
\toprule
\multirow{3}{*}{Dataset}      & \multicolumn{1}{c}{\multirow{3}{*}{Setting}} & \multirow{3}{*}{\begin{tabular}[c]{@{}c@{}}Rectified Noise \\ Setting\end{tabular}} & \multirow{3}{*}{\begin{tabular}[c]{@{}c@{}}Extra \\ SiT Block\end{tabular}} & \multirow{3}{*}{\begin{tabular}[c]{@{}c@{}}Ratio of Added\\  Parameters\end{tabular}} & \multicolumn{5}{c}{\multirow{2}{*}{Metrics}}                                              \\
                              & \multicolumn{1}{c}{}                         &                                                                                     &                                                                             &                                                                                       & \multicolumn{5}{c}{}                                                                      \\ \cline{6-10} 
                              & \multicolumn{1}{c}{}                         &                                                                                     &                                                                             &                                                                                       & FID $\downarrow$ & IS $\uparrow$ & sFID $\downarrow$ & Prec. $\uparrow$ & Rec. $\uparrow$ \\ \hline
\multirow{11}{*}{ImageNet-1k} & SiT-XL/2                                     & -                                                                                   &                                                                             -                                                                                   &                                                                                       -                                                                                   &                  10.16&               123.86&                   12.02&                  0.50&                 \textbf{0.62}\\ \cline{2-10} 
                              & \multirow{8}{*}{+  $\Delta$ RN}& \multirow{5}{*}{$\mathcal{N}\left(\mathbf{0},\boldsymbol{\sum}\right)$}                                                               & 0                                                                           &                                                                                       0.39\%&                  9.72&               122.21&                   12.02&                  0.51&                 0.61\\
                              &                                              &                                                                                     & 1                                                                           &                                                                                       3.93\%&                  9.85&               124.40&                   11.63&                  0.51&                 0.61\\
                              &                                              &                                                                                     & 2                                                                           &                                                                                       7.48\%&                  9.75&               130.21&                   11.28&                  0.52&                 0.61\\
                              &                                              &                                                                                     & 4                                                                           &                                                                                       14.56\%&                  9.60&               131.19&                   11.18&                  \textbf{0.53}&                 0.62\\ \cline{3-10} 
                              &                                              & \multirow{5}{*}{$\mathcal{N}\left(\boldsymbol{\mu},\boldsymbol{\sum}\right)$}                                              & 0                                                                           &                                                                                       0.39\%
&                  9.06&               130.21&                   \textbf{11.18}&                  {0.52}&                 0.61
\\
                              &                                              &                                                                                     & 1                                                                           &                                                                                       3.93\%
&                  \textbf{9.05}&               \textbf{132.10}&                   11.23&                  {0.52}&                 \textbf{0.62}
\\
                              &                                              &                                                                                     & 2                                                                           &                                                                                       7.48\%
&                  9.08&               129.58&                   11.31&                  {0.52}&                 \textbf{0.62}
\\
                              &                                              &                                                                                     & 4                                                                           &                                                                                       14.56\%&                  9.15&               131.43&                   11.26&                  {0.52}&                 \textbf{0.62}\\ \hline
\multirow{9}{*}{AFHQ}         & SiT-B/2                                      & -                                                                                   &                                                                             -                                                                                   &                                                                                       -                                                                                   &                  12.33&               9.99&                   28.14&                  0.55&                 0.53\\ \cline{2-10} 
                              & \multirow{6}{*}{+  $\Delta$ RN}& \multirow{4}{*}{$\mathcal{N}\left(\mathbf{0},\boldsymbol{\sum}\right)$}                                                               & 0                                                                           &                                                                                       0.93\%&                  12.20&              \textbf{10.13}&                   28.19&                  0.56&                 \textbf{0.54}\\
                              &                                              &                                                                                     & 1                                                                           &                                                                                       9.17\%&                  11.98&               10.06&                   27.99&                  0.55&                 \textbf{0.54}\\
                              &                                              &                                                                                     & 2                                                                           &                                                                                       17.41\%&                  12.03&               10.01&                   28.10&                  0.55&                 \textbf{0.54}\\ \cline{3-10} 
                              &                                              & \multirow{4}{*}{$\mathcal{N}\left(\boldsymbol{\mu},\boldsymbol{\sum}\right)$}                                              & 0                                                                           &                                                                                       0.93\%
&                  10.62&              \textbf{10.13}&                   26.68&                  0.57&                 \textbf{0.54}\\
                              &                                              &                                                                                     & 1                                                                           &                                                                                       9.17\%
&                  10.52&               9.88&                   \textbf{26.32}&                  0.57&                 0.52\\
                              &                                              &                                                                                     & 2                                                                           &                                                                                       17.41\%&                 \textbf{10.44}&               9.80&                   26.41&                  \textbf{0.58}&                 0.52\\ \hline
\multirow{9}{*}{CelebA-HQ}    & SiT-B/2                                      & -                                                                                   &                                                                             -                                                                                   &                                                                                       -                                                                                   &                  11.25&               3.55&                   18.31&                  0.62&                 0.47\\ \cline{2-10} 
                              & \multirow{6}{*}{+  $\Delta$ RN}& \multirow{4}{*}{$\mathcal{N}\left(\mathbf{0},\boldsymbol{\sum}\right)$}                                                               & 0                                                                           &                                                                                       0.93\%&                  11.18&               \textbf{3.55}&                   18.27&                  0.62&                 \textbf{0.48}\\
                              &                                              &                                                                                     & 1                                                                           &                                                                                       9.17\%
&                  11.16&               3.54&                   18.20&                  0.62&                 \textbf{0.48}\\
                              &                                              &                                                                                     & 2                                                                           &                                                                                       17.41\%&                  11.15&               3.54&                   18.26&                  0.62&                 \textbf{0.48}\\ \cline{3-10} 
                              &                                              & \multirow{4}{*}{$\mathcal{N}\left(\boldsymbol{\mu},\boldsymbol{\sum}\right)$}                                              & 0                                                                           &                                                                                       0.93\%
&                  \textbf{7.73}&               3.37&                  \textbf{ 14.73}&                  0.70&                 0.45\\
                              &                                              &                                                                                     & 1                                                                           &                                                                                       9.17\%
&                  7.75&               3.39&                   14.78&                  \textbf{0.71}&                 0.45\\
                              &                                              &                                                                                     & 2                                                                           &                                                                                       17.41\%&                  7.74&               3.38&                   14.74&                  0.70&                 0.45\\
\bottomrule
\end{tabular}
\caption{\textbf{Evaluation of Rectified Noise.} The performance of generative models using Rectified Noise on the different dataset at a resolution of 256x256 without Classifier-Free Guidance (CFG), evaluated under different rectified noise settings. ↑ indicates that higher values are better, with ↓ denoting the opposite.}
\label{result}
\end{table*}

We fine-tune the pre-trained RF model using the strategy shown in Figure \ref{pipe1} (a). We extract the pre-trained RF model's features before the linear layer and use them as input for the $\pi$-noise generator. We stack additional SiT blocks, which then connect to a final linear layer to predict the $\pi$-noise. The linear layer is initialized with zeros to ensure that the initial prediction matches the original RF model output. Algorithm \ref{alg:algorithm2} illustrates the implementation of an arbitrary batch step.
    
\section{Experiments}
In this section, we design experiments to investigate the
following questions:
\begin{enumerate}
    \item[\textbf{Q1}] Does employing $\Delta$RN lead to an improvement in RF model performance?
    \item[\textbf{Q2}] Which reparameterizable distribution is most suitable for modeling the $\pi$-noise distribution?
    \item[\textbf{Q3}] Which optimization strategy is more suitable: simultaneous optimization of $\theta$ and ${\psi}$, or optimizing ${\psi}$ first and then $\theta$?
\end{enumerate}

\subsection{Experimental Setup}
\subsubsection{Implementation Details}
\quad\\
We strictly follow the setup in SiT (Ma et al., 2024a) unless otherwise specified. We use linear interpolation to align with the RF optimization objective. We use ImageNet (1.28 million images, 1,000 categories) \cite{deng2009imagenet}, AFHQ (16,130 images of animal faces, 3 categories) \cite{kim2019u}, and CelebA-HQ (30,000 images of celebrity facial images, 2 categories) \cite{karras2017progressive} as training datasets. The model's input for all datasets is 256x256. Each image is then encoded into a compressed vector $z\in R^{32\times32\times4} $ using the Stable Diffusion VAE \cite{rombach2022high}. For model configurations, we use the B/2 and XL/2 architectures introduced in the DiT papers, which process inputs with a patch size of 2. 

\subsubsection{Evaluation Protocol}
\begin{figure*}[t]
\centering
\includegraphics[width=0.92\textwidth]{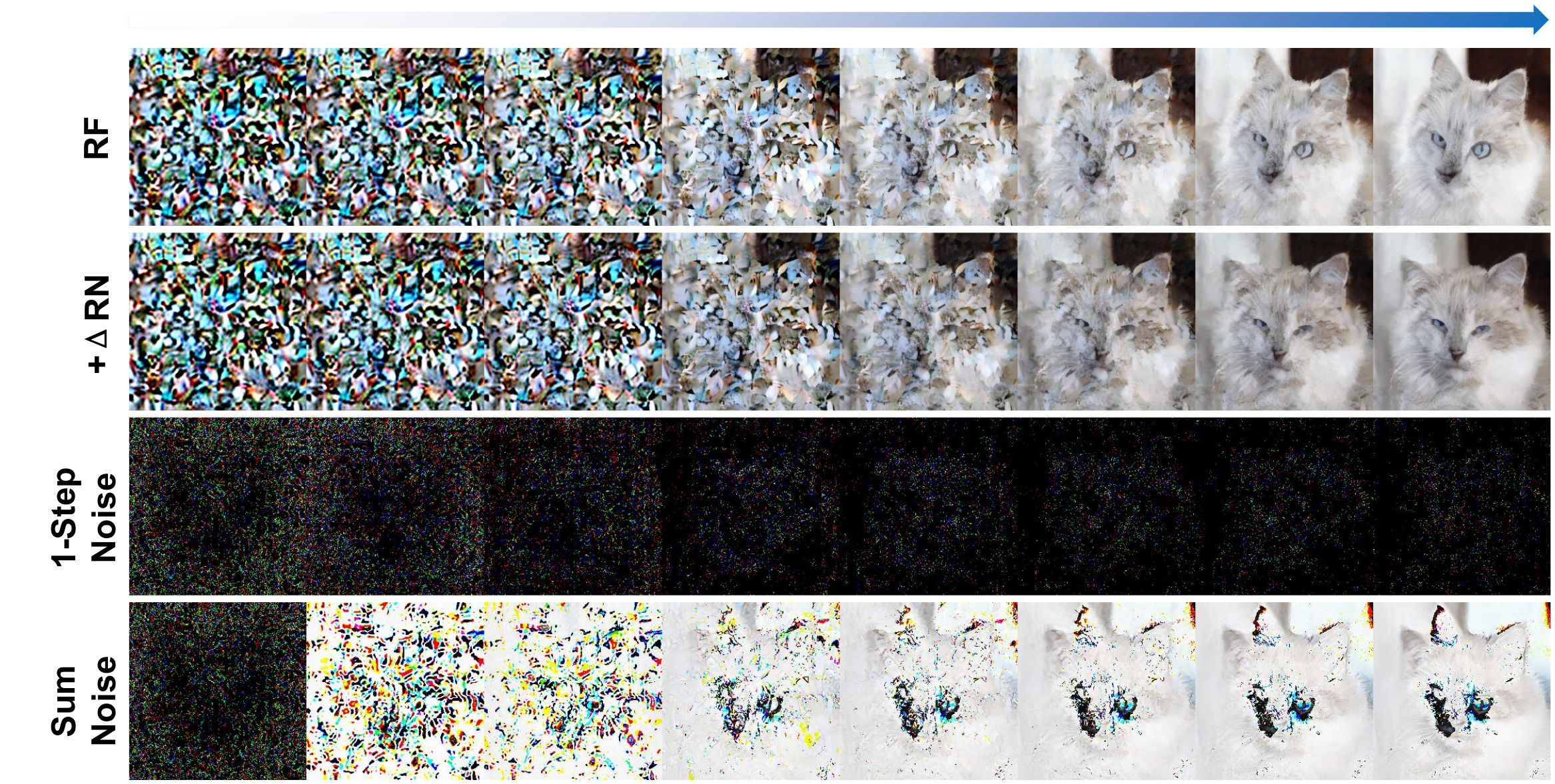}
\caption{\textbf{Visualization of the $\pi$-noise by $\Delta$RN.} The first line shows the original image generation with RF model, the second line shows the results of RF models using $\Delta$RN in one step, the third line shows the generated noise for one step and the fourth line shows the cumulative noise for each time step. We use 180 steps for visualization.}
\label{vision}
\end{figure*}

\quad \\
To comprehensively evaluate image generation quality across multiple dimensions, we employ a rigorous set of quantitative metrics, all computed on a standardized set of generated samples to ensure statistical reliability. For ImageNet, we use 50k generated samples to compute FID for assessing realism, structural FID (sFID) \cite{nash2021generating} for evaluating spatial coherence and Inception Score (IS) \cite{salimans2016improved} for measuring class-conditional diversity, as well as precision (Prec.) for quantifying sample fidelity and recall (Rec.) \cite{kynkaanniemi2019improved} for evaluating coverage of the target distribution. For AFHQ and CelebA-HQ, we generated 15k images and 30k images for evaluation. All evaluations are performed using the SDE Euler–Maruyama solve with 100 steps. The generated images from the standard SiT model and the model using $\Delta$RN are shown in Figure \ref{fig2}. We also visualized the generated $\pi$-noise over time, as shown in Figure \ref{vision}.
\subsection{Rectified Noise Improves SiT}
For the ImageNet dataset, a pre-trained SIT model iterated for 6 million steps was utilized to train $\pi$-noise genartor. The AFHQ dataset used a SIT model pre-trained for 100k steps and the CelebA-HQ dataset used one pre-trained for 200k steps. For both AFHQ and CelebA-HQ datasets, the optimization steps were set to 10k. The results are summarized in Table \ref{result}. 
Overall, $\Delta$RN improves RN in nearly all metrics and all datasets.On the ImageNet, employing $\Delta$RN with SiT-XL/2 lowers FID by up to 1.11. Furthermore, it achieved FID improvements of 1.89 and 3.52 on the AFHQ and CelebA-HQ datasets respectively.

Notably, the number of SiT blocks provides limited performance gains, as a small parameter count is sufficient to achieve good results.

\subsection{Different Noise Analysis}
We employ the fine-tuning strategy to train the $\pi$-noise generator. This is done by building upon a RF model that had been pre-trained for 6 million iterations on ImageNet. We explored three different noise assumptions—Gaussian distribution, Gumbel distribution, and Uniform distribution—training each for 100k iterations. The final results for each metric are summarized in Table \ref{tab:performance}. Uniform, Gumbel, and Gaussian distributions all improve model performance. Among these, Gaussian distribution is the most effective to enhance model performance.
\begin{table}[h!]
\centering
\begin{tabular}{rccccc}
\toprule
 & \multicolumn{5}{c}{Metrics} \\
\cmidrule{2-6}
 Model & FID $\downarrow$ & IS $\uparrow$ & sFID $\downarrow$ & Prec. $\uparrow$ & Rec. $\uparrow$ \\
\midrule
 SiT-XL/2 & 10.16 & 123.86&  12.02& 0.50&  {0.62}\\
Gumbel &9.42&129.73&11.42&0.52&0.61\\
\rowcolor[HTML]{E4E4E4} 
Gaussian &\textbf{9.05}&\textbf{132.10}&\textbf{11.23}&\textbf{0.52}&\textbf{0.62}
\\
Uniform &10.02&124.40&11.63&0.51&0.62\\

\bottomrule
\end{tabular}
\caption{\textbf{ImageNet-1k (256x256)
results of different noise assumptions}. Gaussian noise is the most effective in enhancing the performance of the SiT model.}
\label{tab:performance}
\end{table}
\subsection{Different Training Strategies}
\begin{figure}[h]
\centering
\includegraphics[width=0.48\textwidth]{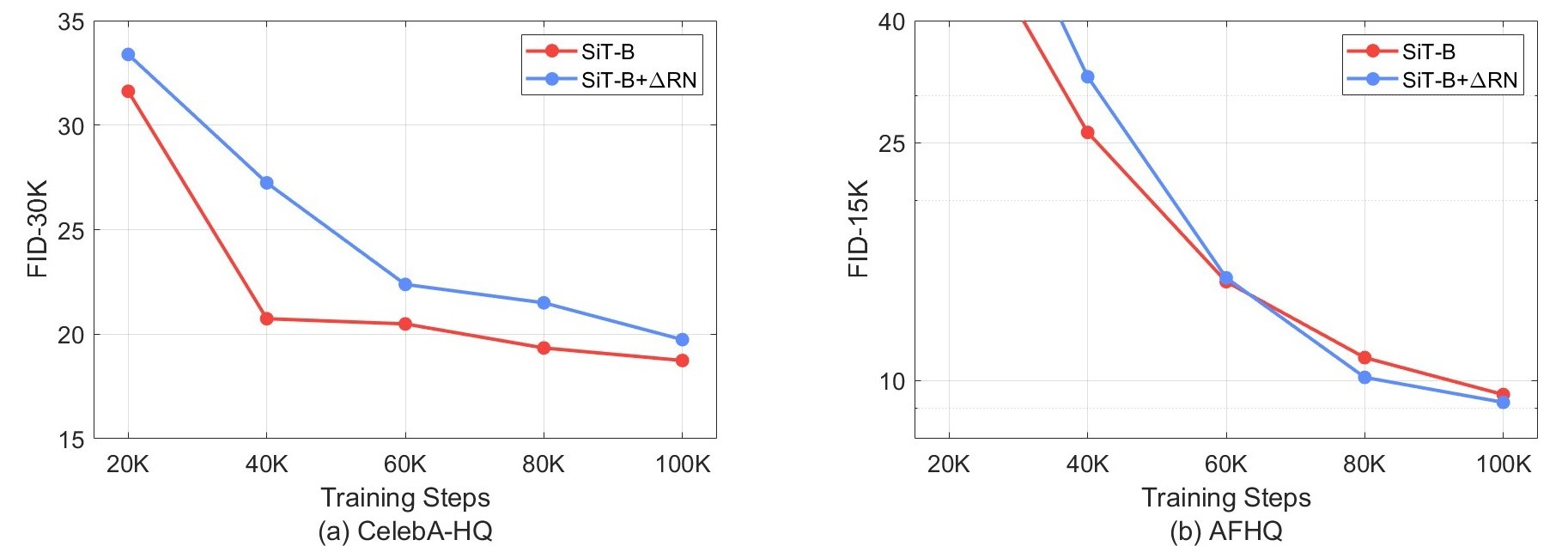}
\caption{{Training FID comparison for SiT-B/2 and SiT-B/2+$\Delta$RN.} The SiT B/2 + $\Delta$RN model converges slower than the SiT B/2 model.}
\label{fid}
\end{figure}

We train the $\Delta$RN model using a strategy that simultaneously optimizes the parameters $\theta$ and ${\psi}$
. On the AFHQ and CelebA-HQ datasets, the trends in FID scores for SiT-B/2 and SiT-B/2 + $\Delta$RN are shown in the Figure \ref{fid}.

From the FID trend graph, we can see that training the $\pi$-noise with the traditional RF method during the training process does not yield a notable FID improvement. While theoretically there are two strategies to optimize the $\theta$ of $\pi$-noise generators, introducing random noise during training leads to instability and difficulty converging to an optimal solution. It is more advisable to employ a fine-tuning strategy to train the $\Delta$RN model.

\section{Conclusion}
In this work, we introduced Rectified Noise, a novel generative model that enhances Rectified Flow models by injecting $\pi$-noise into their velocity fields. We introduce an auxiliary Gaussian distribution related to the flow matching loss to define the task entropy, the core of the $\pi$-noise framework. With the definition of Rectified Flow task entropy, we derive the optimization objective for Rectified Noise. We achieve efficient $\pi$-noise generator training through a fine-tuning approach and validated the effectiveness of this method. Furthermore, we compare the impact of different optimization strategies and noise assumptions on the model. We can further explore the potential of combining flow matching with $\pi$-noise in the future work.

\newpage
\appendix
\bibliography{aaai2026}
\newpage

\urlstyle{rm} 
\def\UrlFont{\rm}  

\frenchspacing  
\setlength{\pdfpagewidth}{8.5in} 
\setlength{\pdfpageheight}{11in} 
%

\definecolor{my_red}{RGB}{201, 132, 132}
\definecolor{my_orange}{RGB}{206, 129, 82}
\definecolor{my_yellow}{RGB}{193, 165, 123}
\definecolor{my_green}{RGB}{120, 166, 73}
\definecolor{my_cyan}{RGB}{106, 156, 137}
\definecolor{my_blue}{RGB}{109, 147, 180}
\definecolor{my_purple}{RGB}{133, 134, 200}
\definecolor{my_gray}{RGB}{124, 136, 148}

%

\lstset{%
	basicstyle={\footnotesize\ttfamily},
	numbers=left,numberstyle=\footnotesize,xleftmargin=2em,
	aboveskip=0pt,belowskip=0pt,%
	showstringspaces=false,tabsize=2,breaklines=true}
\floatstyle{ruled}
\newfloat{listing}{tb}{lst}{}
\floatname{listing}{Listing}
%
\pdfinfo{
/TemplateVersion (2026.1)
}

\setcounter{secnumdepth}{2} 

%



\title{Rectified Noise: A Generative Model Using Positive-incentive Noise\\[1ex]
Supplementary Materials 
}



\maketitle

\section{Model Configuration Details.}  
The specific parameters of the SiT model\cite{ma2024sit} used are shown in the Table \ref{tab1}.
\begin{table}[H]

\centering
\begin{tabular}{ccccc}
\toprule
Config&Layers&Hidden dim&Heads&Params(M) \\
\midrule
B/2&12&768&12&130 \\
XL/2&28&1152&16&675 \\
\bottomrule
\end{tabular}
\caption{{Model configuration details}}
\label{tab1}
\end{table}
\section{Proof of the Task Entropy of the Generation Task}  

In Section 4.1, we presented the specific expression for task entropy and provided the corresponding result directly. We will now detail the proof of this process.Let $\sigma=\mathcal{L}(\mathbf{x},t; {\psi^*}),$ 
\begin{equation}
\begin{aligned}
     H(\mathcal{T})&= \mathbb{E}_{\mathbf{x},t}  H(p(\alpha|\mathbf{x},t)) \\
     &= - \mathbb{E}_{\mathbf{x},t} \int_{-\infty}^{\infty} p(\alpha|\mathbf{x},t) \ln p(\alpha|\mathbf{x},t) \,d\alpha\\
&=  -\mathbb{E}_{\mathbf{x},t}\int_{-\infty}^{\infty} p(\alpha|\mathbf{x},t)) \ln\left( \frac{1}{\sqrt{2\pi\sigma^2}} \exp\left(-\frac{\alpha^2}{2\sigma^2}\right) \right)d\alpha\\
 &=    -\mathbb{E}_{\mathbf{x},t}\int_{-\infty}^{\infty}p(\alpha|\mathbf{x},t)\left[ -\ln(\sqrt{2\pi\sigma^2}) - \frac{\alpha^2}{2\sigma^2} \right] \,d\alpha\\
 & =\ln(\sqrt{2\pi\sigma^2})\mathbb{E}_{\mathbf{x},t}\int_{-\infty}^{\infty}p(\alpha|\mathbf{x},t)\,d\alpha+\\
& \quad \frac{1}{2\sigma^2}\mathbb{E}_{\mathbf{x},t}\int_{-\infty}^{\infty}p(\alpha|\mathbf{x},t)\alpha^2d\alpha\\
 &=\ln(\sqrt{2\pi\sigma^2})+\frac{1}{2} \\
    &= \frac{1}{2} \mathbb{E}_{\mathbf{x},t}  \mathcal{L}(\mathbf{x},t; {\psi^*})+\frac{1}{2} \ln(2\pi e).
\end{aligned}
\end{equation}
\section{Evaluation of Conditional Generation}  
Our proposed pipeline is compatible with the Classifier-Free Guidance (CFG). To further validate the model's performance under standard guidance conditions, we conducted conditional generation experiments on the ImageNet dataset, as shown in the Table \ref{tab2}.
\begin{table}[!htbp]
\centering
\begin{tabular}{ccc}
\toprule
Model&Extra SiT Block&FID \\
\midrule
SiT-XL/2&-&2.20 ± 0.029\\
+ $\Delta$ RN&0&2.06 ± 0.026\\
+ $\Delta$ RN&1&2.05 ± 0.021\\
\bottomrule
\end{tabular}
\caption{\textbf{Performance under Classifier-Free Guidance.} The table summarizes the performance of the proposed model on ImageNet using a guidance scale of $\text{cfg}=1.5$. The experiments were conducted using five distinct random seeds. }
\label{tab2}
\end{table}

\section{Computational Efficiency}  
The inference costs are reported in Table \ref{tab3}. The results indicate that utilizing Rectified Noise requires a marginal extra computational cost.
\begin{table}[!htbp]
\centering
\begin{tabular}{cccc}
\toprule
Model&SiT Block&FLOPs(G)&GPU Mem.(GB) \\
\midrule
SiT-XL/2&-&114.42&2.58\\
+ $\Delta$ RN&0&114.84&2.60\\
+ $\Delta$ RN&1&118.52&2.69\\
\bottomrule
\end{tabular}
\caption{\textbf{Inference Computational Cost of $\text{SiT-XL/2}$ with $\Delta \text{RN}$}.
The table details the cost metrics, including FLOPs and GPU Memory usage. }
\label{tab3}
\end{table}
\section{Visualization of the $\pi$-Noise}  
\begin{figure}[!htbp]
\centering
\includegraphics[width=0.45\textwidth]{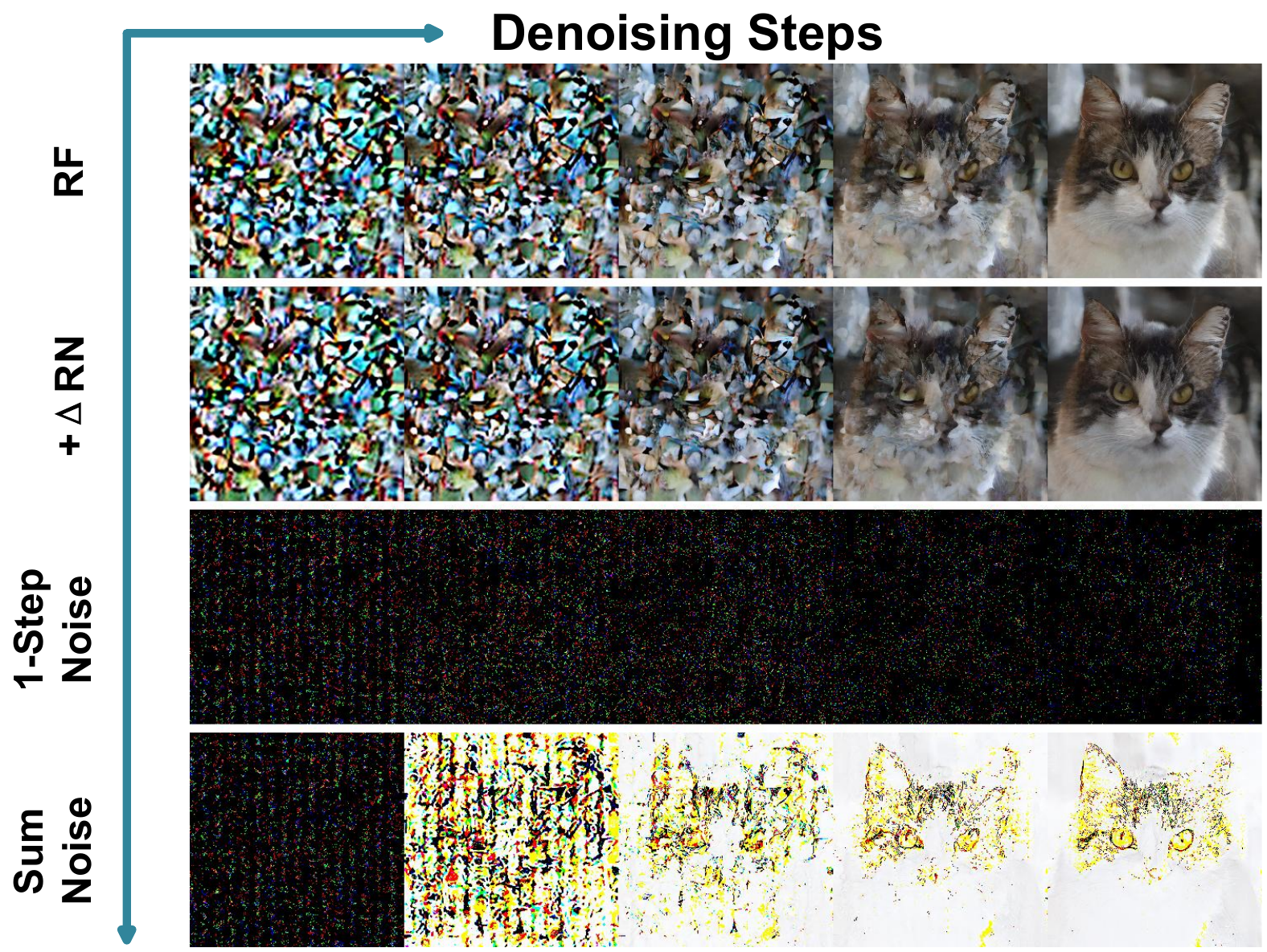}
\caption{\textbf{Visualization of the $\pi$-noise by $\Delta$RN.} The first line shows the original image generation with RF model, the second line shows the results of RF models using $\Delta$RN in one step, the third line shows the generated noise for one step and the fourth line shows the cumulative noise for each time step. We use 180 steps for visualization.}
\label{vision1}
\end{figure}

Sometimes the changes introduced by $\Delta$RN can be difficult to see, they still have an impact on image texture, lighting, and other attributes as shown in Figure \ref{vision1}.


\end{document}